\pdfoutput=1
\documentclass[11pt]{article}

\usepackage{emnlp2021}

\usepackage{times}
\usepackage{latexsym}
\usepackage{booktabs}
\usepackage{colortbl}
\usepackage{multirow}
\usepackage{subcaption}
\usepackage{makecell}

\usepackage[T1]{fontenc}
\usepackage[utf8]{inputenc}

\usepackage{microtype}

\usepackage{graphicx}
\usepackage{amsfonts}
\usepackage{amssymb}
\usepackage{amsmath}
\usepackage{algorithm}
\usepackage{xcolor}
\usepackage[noend]{algpseudocode}

\newcommand{\vecb}[1]{\mathbf{#1}}

\title{A Scalable Framework for Learning From Implicit User Feedback to Improve Natural Language Understanding in Large-Scale Conversational AI Systems}

 \author{Sunghyun Park\thanks{\quad Equal contribution.}\,, Han Li\textsuperscript{\rm *}, Ameen Patel, Sidharth Mudgal, Sungjin Lee, Young-Bum Kim, \\ \bf{Spyros
Matsoukas, Ruhi Sarikaya}\\
 Amazon Alexa AI \\
  \small{\texttt{\{sunghyu, lahl, paameen, sidmsk, sungjinl, youngbum, matsouka, rsarikay\}@amazon.com}}}

\begin{document}
\maketitle
\begin{abstract}
Natural Language Understanding (NLU) is an established component within a conversational AI or digital assistant system, and it is responsible for producing semantic understanding of a user request. We propose a scalable and automatic approach for improving NLU in a large-scale conversational AI system by leveraging implicit user feedback, with an insight that user interaction data and dialog context have rich information embedded from which user satisfaction and intention can be inferred. In particular, we propose a domain-agnostic framework for curating new supervision data for improving NLU from live production traffic. With an extensive set of experiments, we show the results of applying the framework and improving NLU for a large-scale production system across 10 domains.
\end{abstract}

\section{Introduction}
For a conversational AI or digital assistant system \cite{kepuska2018next}, Natural Language Understanding (NLU) is an established component that produces semantic interpretations of a user request, which typically involves analysis in terms of domain, intent, and slot \cite{el2014extending}. For instance, the request \emph{``Play a song by Taylor Swift"} can be interpreted as falling within the scope of \emph{Music} domain with \emph{Play Song} intent and \emph{Taylor Swift} identified for \emph{Artist} slot.

Without an accurate semantic understanding of the user request, a conversational AI system cannot fulfill the request with a satisfactory response or action. As one of the most upstream components in the runtime workflow \cite{sarikaya2017technology}, NLU's errors also have a wider blast radius that propagate to all subsequent downstream components, such as dialog management, routing logic to back-end applications, and language generation.

A straight-forward way to improve NLU is through human annotations, but they are labor-intensive and expensive. Such annotations require at least multiple tiers of annotations (e.g., end user experience, error attribution, and semantic interpretation), and it is hard to consider all relevant contextual conditions. They are also limited by the existing annotation guidelines that may be outdated or that may not accurately reflect user expectations. Due to these limitations, leveraging user feedback, both implicit and explicit, from real production systems is emerging as a new area of research. 

\begin{figure}[t]
\centering
\includegraphics[width=\columnwidth]{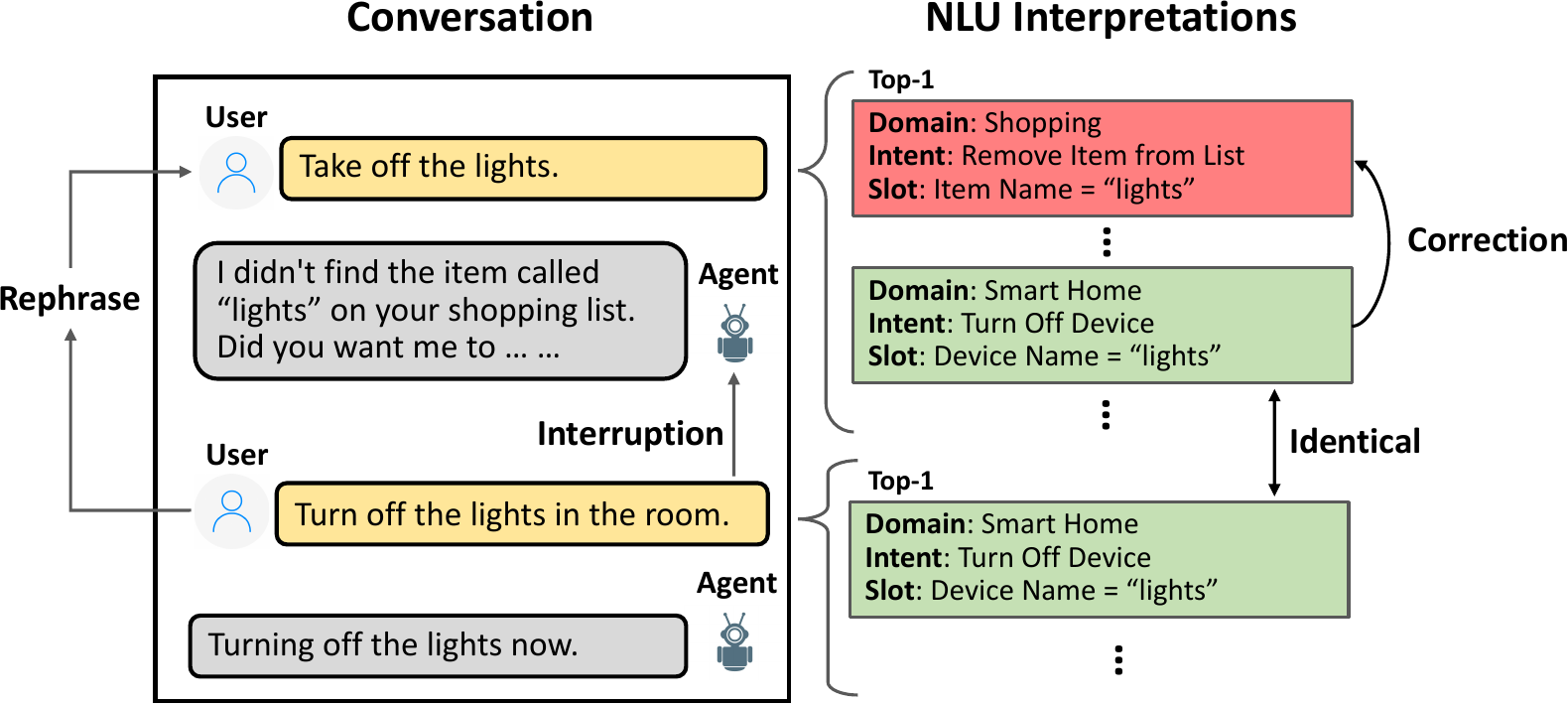}
\caption{An example of implicit user feedback, specifically an indication of user dissatisfaction and user rephrase behavior, that can be used to create new supervision data to correct NLU errors. The left side shows the dialog history and the right side shows the ranked NLU interpretations for each user request. }\label{fig:example_intro}
\end{figure}

Our work makes three main contributions. First, this work is the first in the literature to introduce a scalable, automatic and domain-agnostic approach for leveraging implicit user feedback to continuously and directly improve the NLU component of a large-scale conversational AI system in production. This approach can be applied week over week to continuously and automatically improve NLU towards better end-to-end user experience, and given that no human annotation is required, the approach also raises minimal user privacy concerns. Our approach of using implicit feedback is based on our insight that user interaction data and dialog context have rich information embedded from which user satisfaction and intention can be inferred (see Figure \ref{fig:example_intro}). Second, we propose a general framework for curating supervision data for improving NLU from live traffic that can be leveraged for various subtasks within NLU (e.g., domain/intent classification, slot tagging, or cross-domain ranking). Last, we show with an extensive set of experiments on live traffic the impact of the proposed framework on improving NLU in the production system across 10 widely used domains.

\section{Background and Problem Definition}
\label{sec:related_work}
The NLU component typically has three main types of underlying models - domain classifiers, intent classifiers, and slot taggers \cite{el2014extending}. The three modeling tasks can be treated independently \cite{gao2018neural} or as a joint optimization task \cite{liu2016attention, hakkani2016multi}, and some systems have a model to rank across all domains, intents and slots on a certain unit of semantic interpretation \cite{su2018re}.

Leveraging implicit feedback from the users has been widely studied in the context of recommendation systems \cite{hu2008collaborative, he2016fast, liu2010personalized, loni2018factorization, rendle2012bpr, he2016vbpr, he2016fast, wang2019adversarial} and search engines \cite{joachims2002optimizing, sugiyama2004adaptive, shen2005context, bi2019leverage}. In such systems, common types of implicit user feedback explored include a history of browsing, purchase, click-through behavior, as well as negative feedback.

Leveraging implicit feedback in the context of conversational AI systems is relatively unexplored, but it has been applied for rewriting the request text internally within or post the Automatic Speech Recognition (ASR) component \cite{flare}, improving the Natural Language Generation component \cite{zhang2018exploring}, and using user engagement signals for improving the entity labeling task specifically focused on Music domain \cite{muralidharan2019leveraging}. We note that compared to explicit feedback \cite{petrushkov2018learning,iyer2017learning}, using implicit feedback is more scalable and does not introduce friction in user experience. But it comes with a challenge of the feedback being noisy, and leveraging the feedback is more difficult when there is no sufficient data such as for improving tail cases \cite{wang2021handling,wang2021learning}.

In this paper, we specifically focus on two types of implicit user feedback - \emph{dissatisfaction} of experience (to understand what to fix, e.g., users prematurely interrupting a system's response) and \emph{clarification} of intention through rephrase (to understand how to fix, e.g., users clarifying their requests by rephrasing the previous request in simpler terms). In this work, we assume that there are mechanisms already in place to automatically (1) infer user dissatisfaction ($f_{defect}$ in Section \ref{subsec:prob_def}) and also (2) detect whether a given request is a rephrase of a previous request ($f_{rephrase}$ in Section \ref{sec:sol_framework}). There are many ways to build these two mechanisms, either rule-based or model-based. Due to space limitation, we leave more details of the two mechanisms outside the scope of this paper. For completeness and better context to the reader however, we briefly describe various ways to build them, which would be straight-forward to adapt and implement.

\subsection{User Dissatisfaction Detection}
Unless we specifically solicit users' feedback on satisfaction after an experience, user feedback is mostly implicit. There are many implicit user behavior signals that can help with detecting user dissatisfaction while interacting with a conversational AI system. They include termination (stopping or cancelling a conversation or experience), interruption (barging in while the system is still giving its response), abandonment (leaving a conversation without completing it), error-correcting language (preceding the follow-up turn with "no, ..." or "I said, ..."), negative sentiment language showing frustration, rephrase or request reformulation, and confirmation to execute on an action \cite{Beaver_Mueen_2020,sarikaya2017technology}.

Although not strictly from the user behavior, there are other signals from the system action and response that are also useful. They include generic error-handling system responses ("I don't know that one."), the templates executed for generating natural language error-handling responses (the song entity is not found for playing music), and the absence of a response \cite{Beaver_Mueen_2020,sarikaya2017technology}. There are also component-level signals such as latency or low confidence scores for the underlying models within each component such as ASR or NLU.

For more advanced approaches, we can combine the signals from the user behavior and the system together, try to model user interaction patterns, and use additional context from past interaction history beyond immediate turns \cite{jiang2015automatic,ultes2014interaction,bodigutla2020joint}. Furthermore, user satisfaction can depend on usage scenarios \cite{kiseleva2016understanding}, and for specific experiences like listening to music, we can adapt related concepts such as dwell time in the search and information retrieval fields to further fine-tune.

\subsection{User Rephrase Detection}
There are many lines of work in the literature that are closely related to this task under the topics of text/sentence semantic similarity detection and paraphrase detection. The approaches generally fall into lexical matching methods \cite{manning1999foundations}, leveraging word meaning or concepts with a knowledge base such as WordNet \cite{mihalcea2006corpus}, latent semantic analysis methods \cite{landauer1998introduction}, and those based on word embeddings \cite{camacho2018word} and sentence embeddings \cite{reimers2019sentence}. In terms of modeling architecture, Siamese network is common and has been applied with CNN \cite{hu2014convolutional}, LSTM \cite{mueller2016siamese}, and BERT \cite{reimers2019sentence}. The task is also related to the problems in community question-answering systems for finding semantically similar questions and answers \cite{srba2016comprehensive}.

\subsection{Problem Definition}
\label{subsec:prob_def}
Denote $\mathcal{T} = (\Sigma, \Pi, N, A)$ to be the space of all user interactions with a conversational AI system with each request or turn $t_i = (u_i, p_i, c_i, a_i) \in \mathcal{T}$ consisting of four parts: $u_i \in \Sigma$ is the user request utterance, $p_i \in \Pi$ is the semantic interpretation for $u_i$ from NLU, $c_i \in N$ is the contextual metadata (e.g., whether the device has a screen), and $a_i \in A$ is the system action or response. Here, we are proposing a general framework that allows a scalable and automatic curation of supervision data to improve NLU, and we keep the unit of the semantic interpretation abstract for generalizability, which can be for one or a combination of NLU sub-tasks of domain classification, intent classification, and slot tagging. For instance, one possible interpretation unit would be domain-intent-slots tuple, which is what we use in our experiments described in Section \ref{sec:exp}. Although we only focus on NLU in this paper, the approach here can be extended to improve other components in a conversational AI system such as skill routing \cite{li2021neural}.

We define a \emph{session} of user interaction $s = \{t_1, t_2, \dots, t_q\} \subseteq \mathcal{T}$ which is a list of time-consecutive turns by the same user. Denote $m_t$ to be the NLU component at timestamp $t$. We collect the interaction session data $\mathcal{S}_{live} = \{s_1, s_2, \dots, s_n\}$ from live traffic for a certain period of time $\Delta$ (e.g., one week) starting at time $t$, from which we curate new supervision data to produce $m_{t+\Delta}$ with improved performance. Specifically, given a tool $f_{defect}$ for automatic analysis of user dissatisfaction for each turn, we process $\mathcal{S}_{live}$ to identify all turns that indicate user dissatisfaction, $t_i \in \mathcal{D}_{defect}$, which we call a \emph{defective turn} or simply a \emph{defect}. The key challenges then are how to (1) identify \emph{target defects} which are high-confidence defects that can be targeted by NLU (i.e., there is sufficient disambiguation power within NLU that it can learn to produce different results if given specific supervision) and that are likely causing repeated and systematic dissatisfaction of user experience, and (2) find a likely better interpretation for the \emph{target defects} to change system action or response that leads to user satisfaction.

\section{Solution Framework}
\label{sec:sol_framework}
\begin{figure}[t]
\centering
\includegraphics[width=\columnwidth]{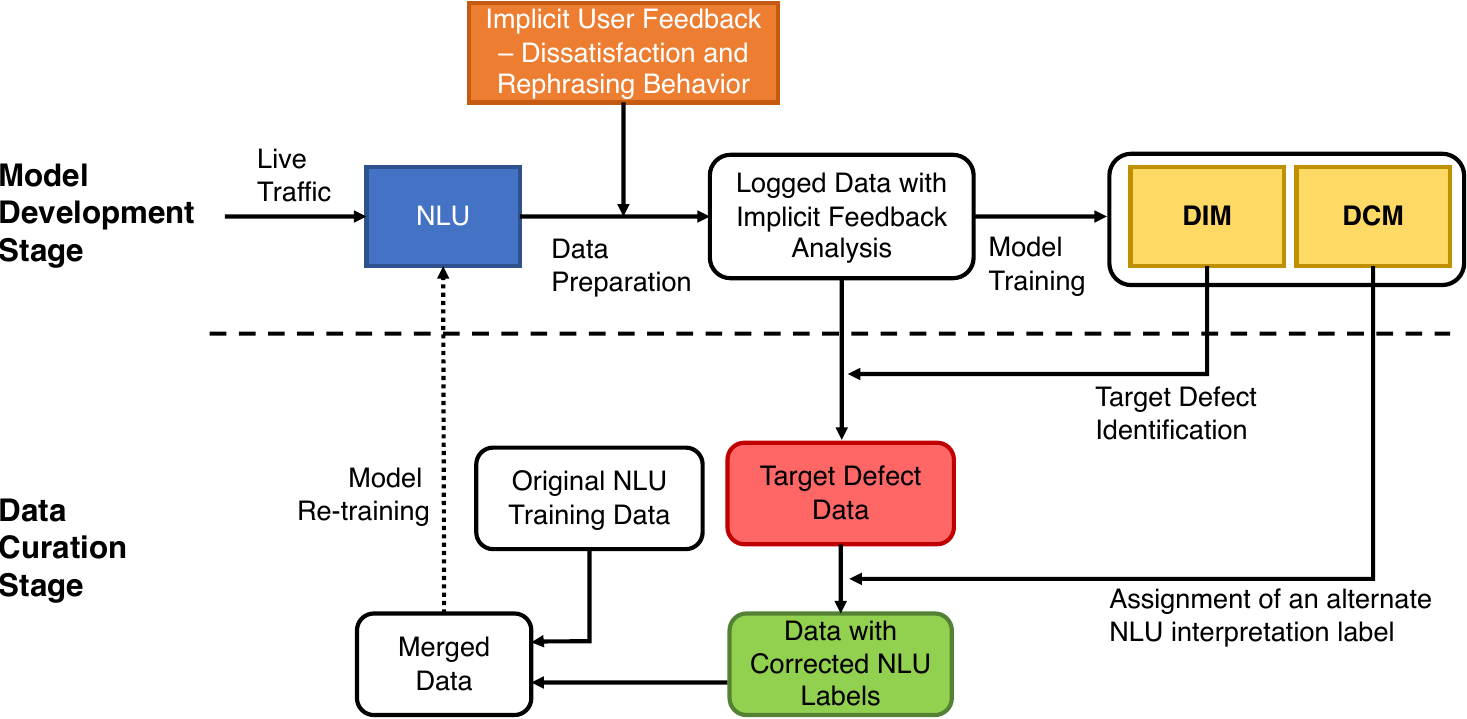}
\caption{Our framework for leveraging implicit user feedback to automatically curate supervision data for improving NLU, consisting of \emph{Defect Identification Model (DIM)} and \emph{Defect Correction Model (DCM).}}\label{fig:framework}
\end{figure}

The framework involves two deep learning models - \emph{Defect Identification Model (DIM)} for addressing the first challenge of identifying \emph{target defects} and \emph{Defect Correction Model (DCM)} for the second challenge of correcting them by automatically labeling them with a likely better semantic interpretation (see Figure \ref{fig:framework}). It is straight-forward to apply DIM and DCM on the production traffic log to curate new supervision data for improving NLU.

\paragraph{Data Preparation:} 
We collect the user interaction session data from the production log $\mathcal{S}_{live}$ for an arbitrary period of time (e.g., past one week). Given a user dissatisfaction analysis tool $f_{defect}$ and a rephrase analysis tool $f_{rephrase}$, we tag $t_j \in s_i$ as a defect if $f_{defect}$ detects user dissatisfaction for the turn and we tag $t_j \in s_i$ as a rephrase if there exists $t_i \in s_i$ where $\emph{j} \textgreater \emph{i}$ (i.e., temporally $t_j$ occurred after $t_i$) and $f_{rephrase}$ detects $t_j$ to be a rephrase of $t_i$. We then extract each turn in $\mathcal{S}_{live}$ to create turn-level data $\mathcal{D}_{live} = \{t_j \in s_i \mid s_i \in \mathcal{S}_{live}\}$ with $t_j$ containing two binary labels of defect $e_d$ and rephrase $e_r$.

\subsection{Defect Identification Model (DIM)}
\label{subsec:dim_train}
We define DIM as $f_{dim}: \mathcal{T} \rightarrow \{0, 1\}$, which takes as input each turn $t_i \in \mathcal{D}_{live}$ and outputs whether $t_i$ is a \emph{target defect} or not. It uses the same contextual features (and architecture) as the underlying individual NLU model we wish to improve and uses the results of $f_{defect}$, or $e_d$, as the ground-truth labels for training. This allows us to filter down the defects into those that can be targeted by the NLU model of interest (since the same features could predict the defects, suggesting enough disambiguation capacity). By tuning the probability threshold used for binary model prediction, we can further reduce noise in defects and focus on more high-confidence defects that are repeated and systematic failures impacting the general user population.

\begin{figure}[t]
\centering
\includegraphics[width=\columnwidth]{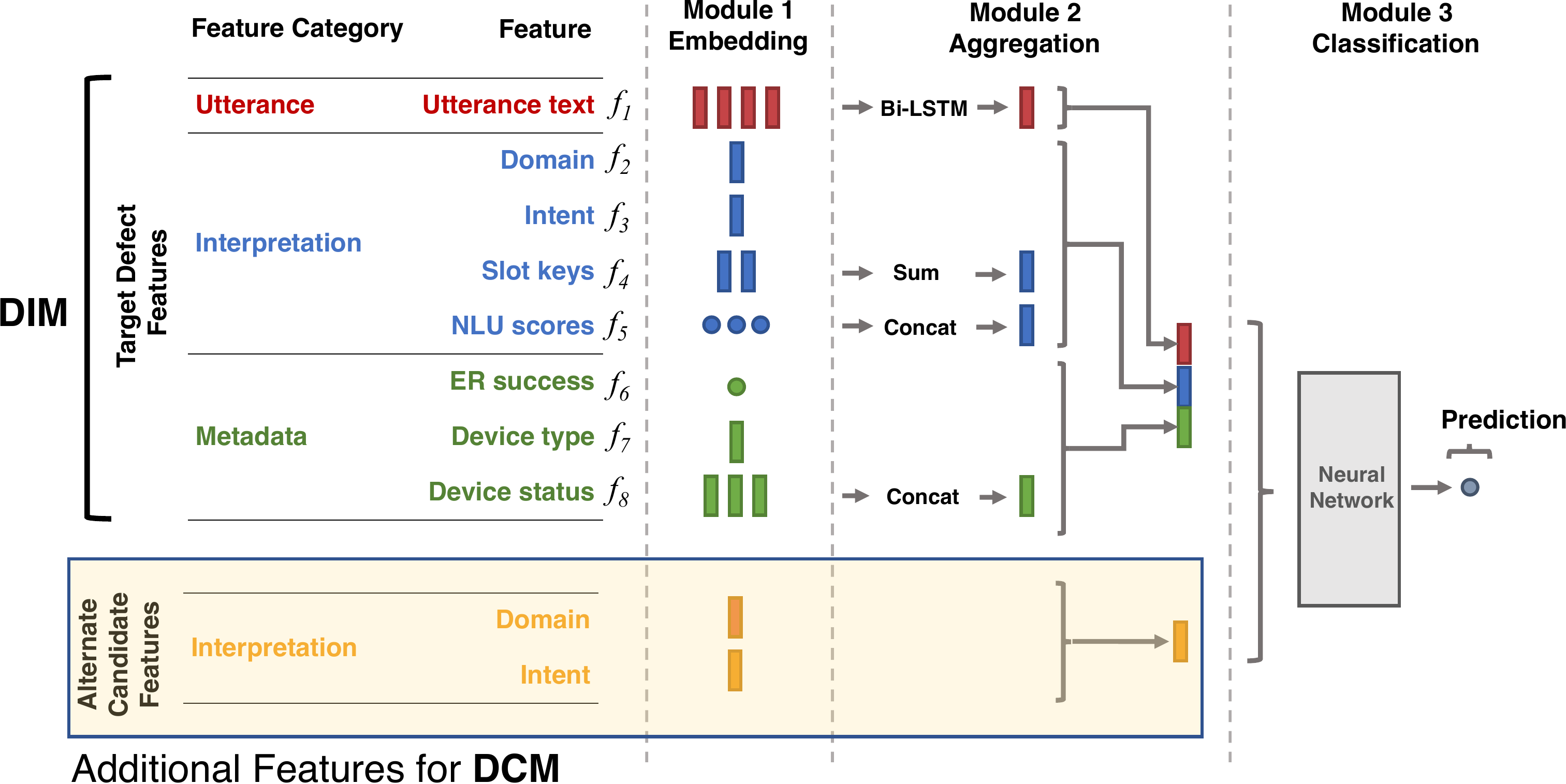}
\caption{The model architectures for Defect Identification Model (DIM) and Defect Correction Model (DCM). For DIM, the prediction is for target defect probability, and for DCM, it is for correction probability (i.e., whether the alternate domain and intent is a good alternate ground-truth label).}\label{fig:dim}
\end{figure}

Figure \ref{fig:dim} shows an example DIM architecture for a cross-domain interpretation re-ranking model (more detail in \ref{subsec:exp_method}). The model architecture consists of three main modules: \emph{embedding}, \emph{aggregation}, and \emph{classification}. Given each feature $f_j$ extracted from $t_i$, the embedding module $H_{emb}$ converts $f_j$ into an embedding. For each sequential or categorical feature $f_j$, denoting $\vecb{w}_{f_j, t_i}$ as the value of $f_j$ with $m$ tokens (where $m$=1 for categorical), we generate $\vecb{v}_{f_j, t_i} = H_{emb}(\vecb{w}_{f_j, t_i}) \in \mathbb{R}^{m \times d_{f_j}}$ with each token converted into the $d_{f_j}$-dimensional embedding. For each numerical feature, we have $\vecb{v}_{f_j, t_i} = \vecb{w}_{f_j, t_i}$ as each feature is already represented by numeric values. The aggregation module $H_{agg}$ then converts $\vecb{v}_{f_j, t_i}$ of each feature $f_j$ to an aggregation vector $\vecb{u}_{f_j, t_i}$ that summarizes the information of $\vecb{v}_{f_j, t_i}$. Based on the feature type, $H_{agg}$ applies different aggregation operations. For example, we apply a Bi-LSTM \cite{schuster1997bidirectional} to the utterance text embeddings $\vecb{v}_{f_1, t_i}$ to capture the word context information. Finally, the classification module $H_{cls}$ takes as input all aggregation vectors to make a prediction whether $t_i$ is a target defect or not. Specifically, we first concatenate all aggregation vectors to get a summarization vector $\vecb{u}_{t_i} = \bigoplus_{f_j}\vecb{u}_{f_j, t_i}$. Then, a two-layer highway network \cite{SrivastavaGS15} is applied to $\vecb{u}_{t_i}$ to make a binary prediction. The model is trained using binary cross-entropy loss.

When developing DIM, we split $\mathcal{D}_{live}$ into the training set $\mathcal{D}_{train}$ and the validation set $\mathcal{D}_{valid}$ with a ratio of 9:1. Once we have DIM trained with $\mathcal{D}_{train}$, we use $\mathcal{D}_{valid}$ to further tune the prediction probability threshold used to extract target defects from all defects tagged by $f_{defect}$. Specifically, for each turn $t_i \in \mathcal{D}_{defect}$, we pass it to $f_{dim}$ to get the confidence score $o_i = f_{dim}(t_i)$ of being a defect. Then, we generate the target defect set $\mathcal{D}_{target} = \{t_i \mid o_i > \tau\}$, i.e., we collect all turns satisfying the defect prediction confidence being greater than a threshold $\tau$. In order to select the value for $\tau$, we perform a binary search on $\mathcal{D}_{valid}$ as shown in Algorithm \ref{alg:thres_search}, which takes as inputs two additional parameters $\lambda$ (to set the minimum prediction accuracy we want) and $\epsilon$.

\begin{algorithm}[t]
\small
\caption{DIM threshold determination.}
\begin{algorithmic}
\Procedure{ThresSearch}{$f_{dim}$, $\mathcal{D}_{valid}$, $\lambda$, $\epsilon$}
\State low, high $\gets$ 0, 1
\While{$|$ low - high $| > \epsilon$}
    \State $\tau \gets$ (low $+$ high) $/ 2$
    \State $\mathcal{P}_{valid} \gets \{t_i \mid f_{dim}(t_i) > \tau, \forall t_i \in \mathcal{D}_{valid}\}$
    \State $\alpha \gets$ \textsc{PredictionAccuracy}($\mathcal{P}_{valid}$)
    \If{$\alpha < \lambda$} low $\gets \tau$
    \Else \hspace{1mm} high $\gets \tau$
    \EndIf
\EndWhile\\
\,\,\,\quad\Return{$\tau$}
\EndProcedure
\end{algorithmic}
\label{alg:thres_search}
\end{algorithm}

\subsection{Defect Correction Model (DCM)}
\label{subsec:dcm_train}
We define DCM as $f_{dcm}: \mathcal{T} \times \Pi \rightarrow \{0, 1\}$, which takes as input a pair $(t_i, p_j)$ with $t_i \in \mathcal{D}_{live}$ and $p_j \in \Pi$ to make a prediction whether $p_j$ is a proper semantic interpretation for $t_i$. As the space of the semantic interpretation $\Pi$ is too large, we can make the process more efficient by restricting to find a better interpretation in the $k$-best predictions $P_i^k \subseteq \Pi$ (i.e., $k$ interpretations with the highest prediction confidence) by the NLU model of interest. Note that it is not difficult to force more diversity into the $k$-best predictions by only allowing top predictions from each domain or intent. For training, we leverage rephrase information from the logged data to automatically assign a corrected semantic interpretation as the new ground-truth label for the defects, with the following assumption: Given a pair of turns $t_i$ and $t_j$, if (a) the utterance of $t_j$ rephrases the utterance of $t_i$ in the same session and (b) $t_j$ is non-defective, then the semantic interpretation of $t_j$ is also the correct interpretation for $t_i$.

Following the example DIM architecture for the cross-domain interpretation re-ranking model in Figure \ref{fig:dim}, the DCM architecture extends that of DIM with the main difference that we can generate other features based on domain, intent and slot information from $p_j$. To obtain the training data, we first examine all turns in $\mathcal{D}_{live}$ to generate the high value set $\mathcal{D}_{h} \subseteq \mathcal{T} \times \mathcal{T}$. Each instance $(t_i, r_i) \in \mathcal{D}_{h}$ is a pair of turns satisfying (a) $t_i \in \mathcal{D}_{live}$ is a defect and (b) $r_i \in \mathcal{D}_{live}$ is a non-defective rephrase of $t_i$ in the same session (defects and rephrases are described in Section \ref{subsec:prob_def} and Section \ref{sec:sol_framework}:Data Preparation). We then generate the training data $\mathcal{D}_{train}$ using the high value set $\mathcal{D}_h$. Specifically, for each pair $(t_i, r_i) \in \mathcal{D}_h$, we generate $k$ training instances as follows. First, we get the $k$-best interpretations $P_{r_i}^k$ of $r_i$. Then, we pair $t_i$ with each candidate $p_j \in P_{r_i}^k$ to get a list of tuples $(t_i, p_1), (t_i, p_2), \dots, (t_i, p_k)$. Next, we expand each tuple $(t_i, p_j)$ by assigning a label $c$ indicating whether $p_j$ can be a proper interpretation for $t_i$. Denote $p^* \in P_{r_i}^k$ as the correct interpretation for $r_i$, assumed since it is executed without a defect (note that the top-1 interpretation is not necessarily the executed and correct one, although it is most of the time). We generate one positive instance $(t_i, p^*, c=1)$, and $k-1$ negative instances $\{(t_i, p_j, c=0) \mid p_j \in P_{r_i}^k \land p_j \ne p^*)\}$. Only using the $k$-best interpretations from $r_i$ to generate $\mathcal{D}_{train}$ may not be sufficient, as in practice the value $k$ is small and many interpretations observed in real traffic does not appear in the training data. To make the model generalize better, we augment the training data by injecting random noise. For each pair $(t_i, r_i) \in \mathcal{D}_h$, in addition to the $k-1$ generated negative instances, we randomly draw $q$ interpretations $P_{noise}^q = \{p^n_1, p^n_2, \dots, p^n_q\} \subseteq \Pi$ that are not in $P_{r_i}^k$, and we generate $q$ new negative instances $\{(t_i, p_j^n, c=0) \mid p_j^n \in P_{noise}^q\}$. In short, DCM's role is to find the most promising alternate interpretation in $t_i$'s k-best interpretation list given that $t_i$ is a defect.

\paragraph{New Supervision Data Curation:} Once we have $f_{dcm}$ trained, the last step of the framework is to curate new supervision data by applying $f_{dcm}$ to each turn $t_i \in \mathcal{D}_{target}$ identified by $f_{dim}$ and automatically assigning a better semantic interpretation for correction. Specifically, we pair each turn $t_i \in \mathcal{D}_{target}$ with every interpretation candidate $p_j \in P_i^k$ as the input to $f_{dcm}$. The interpretation with the highest score $p^* = \arg\max_{p_j \in P_{i}^k} f_{dcm}(t_i, p_j)$ is used as the corrected interpretation for $t_i$.

\section{Experiment Results and Discussion}
\label{sec:exp}
% -------> Latex based table ------
\begin{table}
\centering\small
\includegraphics[width=\columnwidth]{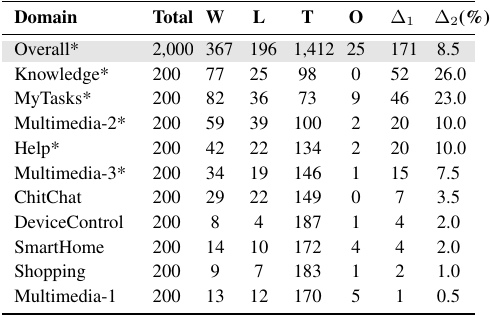}
\caption{Overall side-by-side win-loss evaluation results across 10 domains, comparing the top interpretation prediction between the baseline NLU and the updated NLU improved with our framework. ``W," ``L," ``T" and "O" represent "Win," "Loss," "Tie" and "Others" respectively. A win means that the updated NLU produced a better top interpretation than the baseline (* denotes statistical significance at p\textless.05).}\label{table:overall_eval}
\end{table}

\begin{table*}[ht]
\centering
\includegraphics[width=\textwidth]{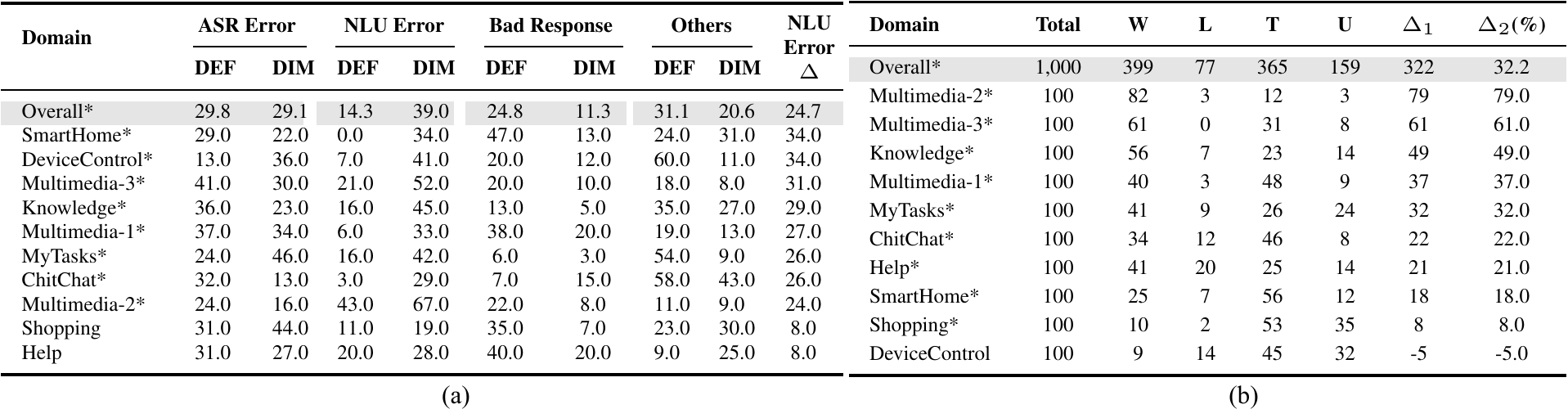}
\caption{(a) The analysis of DIM through error attribution annotations between the defects in the production traffic vs. the target defects identified by DIM. The numbers are in percentage. (b) The analysis of DCM through win-loss annotations between the top interpretation produced by the baseline NLU and the new interpretation label assigned by DCM. Statistical significance at p\textless.05 is noted with *, specifically on the NLU errors in (a).}\label{table:dim_dcm_eval}
\end{table*}

\subsection{Experiment Methodology}
\label{subsec:exp_method}

\begin{figure}[t]
\centering
\includegraphics[width=\columnwidth]{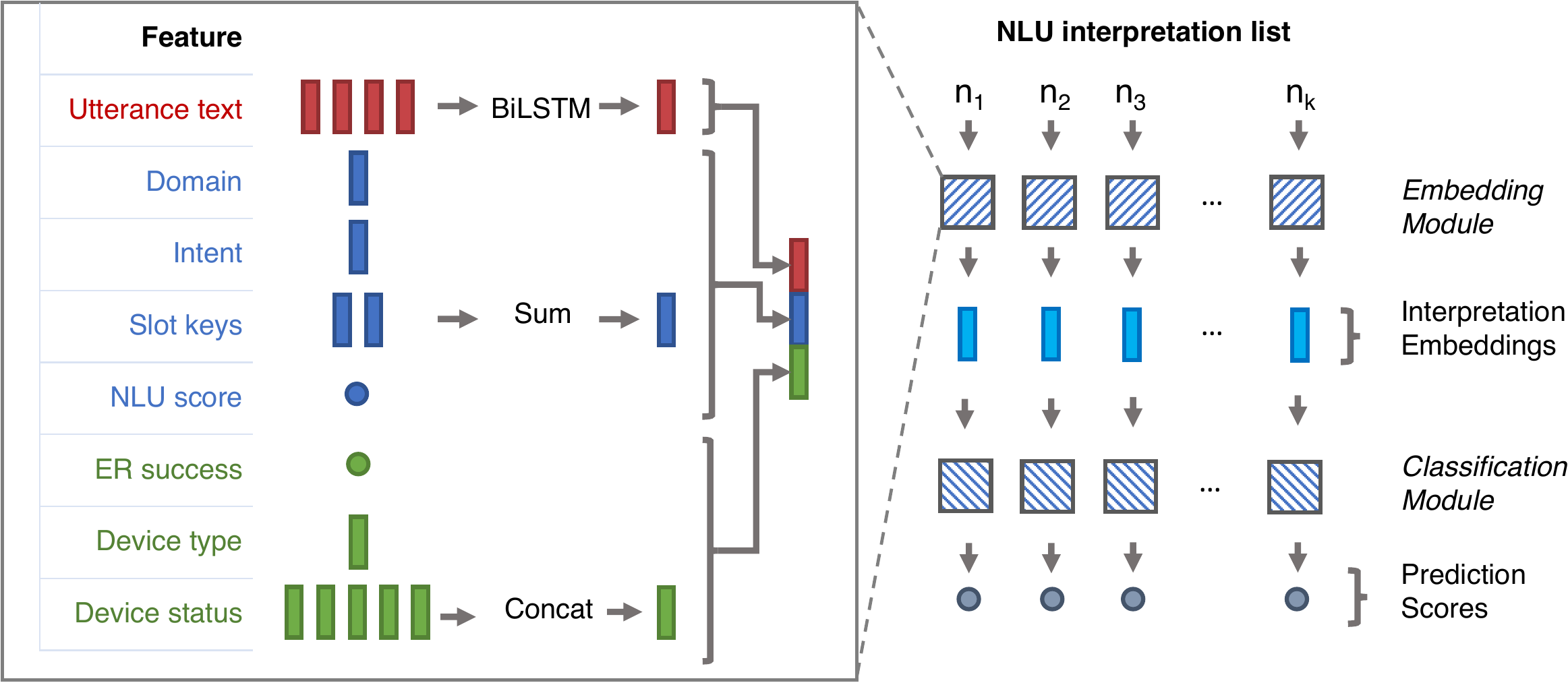}
\caption{The model architecture for the re-ranker, which is a subtask we put at the last layer of the NLU to produce a better ranked list of interpretations.}\label{fig:hyprank}
\end{figure}

\paragraph{Dataset and Experiment Settings:}
Given a baseline NLU in production, $m_{base}$, which produces a ranked list of interpretations with each interpretation comprising domain-intent-slots tuple, we inject a re-ranking subtask at the very last layer of the NLU workflow to build an improved NLU, $m_{new}$. We call the subtask re-ranking because it takes in an already ranked list (i.e., the output of $m_{base}$) and makes a final adjustment. We leverage the new supervision data obtained through our framework to train the re-ranking model for improving the overall NLU performance. Figure \ref{fig:hyprank} shows the model architecture of the re-ranker, which is a simple extension of the DIM architecture, and it learns from the new supervision data when to top-rank a better interpretation that is not at the top of the list (trained with sigmoid activation functions at the output layer and binary cross-entropy loss). We note here that the specific model architecture is not as important as the new supervision data obtained through our framework that is the key for bringing NLU improvements. This experiment setup is appealing in that it is straightforward and simple, especially in the production setting. First, NLU consists of many domain-specific models that are spread out to multiple teams, making it difficult to coordinate leveraging the new supervision data for improvement across multiple domains. Second, working with the final re-ranking model allows us to improve NLU performance domain-agnostically without needing to know the implementation details of each domain. Third, it is easier to control the influence of the new supervision data since we need to manage only one re-ranking component. 

Given sampled and de-identified production traffic data from one time period $\mathcal{D}_{period1}$, which have been analyzed by $f_{defect}$ and $f_{rephrase}$\footnote{In today's production system, $f_{defect}$ and $f_{rephrase}$ show $F_1$ scores over 0.70.}, we first train DIM according to Section \ref{subsec:dim_train}, with over 100MM training instances from $\mathcal{D}_{period1}$ and over 10MM defects identified by $f_{defect}$. Then, we extract over 8MM high-value rephrase pairs (a defective turn and non-defective rephrase in the same session) from $\mathcal{D}_{period1}$ to train DCM according to Section \ref{subsec:dcm_train}. To train the re-ranker, we randomly sample over 10MM instances $\mathcal{D}_{s} \subseteq \mathcal{D}_{period1}$ and over 1MM defects identified by $f_{defect}$. We apply the trained DIM to the sampled defects $\mathcal{F}_{def}$ that filters them down from over 1MM defects to over 300K target defects $\mathcal{F}_{dim}$ that the NLU re-ranker has sufficient features to target and produce different results. Then, all target defects $\mathcal{F}_{dim}$ are assigned a new ground-truth interpretation label by the trained DCM (note that not all defects have corresponding non-defect rephrases, hence the value of DCM for finding the most promising alternate interpretation from the ranked list), which serve as the new curated supervision for building $m_{new}$, while the rest of the non-defective instances keep the top-ranked interpretation as the ground-truth label. In other words, most of the instances in $\mathcal{D}_{s}$ are used to replicate the $m_{base}$ results (a pass-through where the same input ranked list is outputted without any change), except for over 300K (over 3\% of the total training data) that are used to revert the ranking and put a better interpretation at the top.

\emph{Overall Side-by-Side Evaluation}: The overall performance between $m_{base}$ and $m_{new}$ was compared on another sampled production traffic from non-overlapping time period $\mathcal{D}_{period2}$ in a shadow evaluation setting, in which the traffic flowing through $m_{base}$ was duplicated and simultaneously sent to $m_{new}$ that is deployed to the same production setting as $m_{base}$ but without end-user impact. Both $m_{base}$ and $m_{new}$ produced the same ranked list of interpretations for over 99\% of the time. Note that this is by design since incremental improvements are preferred in production systems without drastically changing the system behavior and that our approach can be applied continuously, week over week (changing the proportion of the new supervision data will have an impact on the replication rate). Furthermore, even 1\% change in the overall system behavior has a huge impact at the scale of tens of million of requests per week in a large-scale production system. We performed win-loss annotations on the deltas (when $m_{base}$ and $m_{new}$ produced different results) with in-house expert annotators who follow an established NLU annotation guideline to make a side-by-side evaluation whether $m_{new}$ produced a better interpretation (i.e., win) on the top compared to $m_{base}$ or not (N = 12, agreement = 80.3\%, Cohen’s kappa = 0.60 indicating moderate agreement; note that the annotators are trained to reach agreement level that is practical given the high complexity of the NLU ontology). We randomly sampled 200 such requests per domain that produced different results\footnote{A/B testing results on around 20 intents with over 100MM live utterances showed improvement in reducing defect ratio (i.e., the ratio of utterances tagged by $f_{defect}$) end-to-end from 72.9\% to 42.2\% on the deltas (statistically significant at p\textless.05).}.

\emph{DIM Analysis}: We randomly sampled 100 defects per domain from $\mathcal{F}_{def}$ and $\mathcal{F}_{dim}$ respectively and performed error attribution annotations (i.e., \emph{ASR error} for mis-transcribing \emph{``play old town road"} to \emph{``put hotel road"}, \emph{NLU error} for mis-interpreting \emph{``how do I find a good Italian restaurant around here"} to \emph{Question Answering} intent instead of \emph{Find Restaurant} intent, \emph{Bad Response} for having a correct interpretation that still failed to deliver a satisfactory response or action, and \emph{Others} for those that the annotators could not determine due to lack of context or additional information; N = 12, agreement = 71.3\%, Cohen’s kappa = 0.63 indicating substantial agreement).

\emph{DCM Analysis}: We perform the same win-loss annotations as described in overall shadow evaluation on 100 random samples per domain, specifically on the curated supervision data $\mathcal{F}_{dim}$ with new ground-truth assigned by DCM.

\paragraph{Training Setup:} All the models were implemented in PyTorch \cite{paszke2019pytorch} and trained and evaluated on AWS p3.8xlarge instances with Intel Xeon E5-2686 CPUs, 244GB memory, and 4 NVIDIA Tesla V100 GPUs. We used Adam \cite{kingma2014adam} for training optimization, and all the models were trained for 10 epochs with a 4096 batch size. All three models have around 12MM trainable parameters and took around 5 hours to train.

\subsection{Results and Discussions}

\paragraph{Overall Side-by-Side Evaluation:} Table \ref{table:overall_eval} shows the overall shadow evaluation results, making NLU-level comparison between $m_{base}$ and $m_{new}$. The column \emph{Total} shows the number of requests annotated per domain. The columns \emph{Win}, \emph{Loss}, and \emph{Tie} show the number of requests where $m_{new}$ produced better, worse, and comparable NLU interpretations than $m_{base}$ respectively. The column \emph{Others} shows the number of requests where the annotators could not make the decision due to lack of context. The column $\Delta_1$ shows the difference in the number of win and loss cases, and $\Delta_2$ shows the relative improvement (i.e., $\Delta_1$ / \emph{Total} in percentage). First, we note that $m_{new}$ overall produced a better NLU interpretation on 367 cases while making 196 losses, resulting in 171 absolute gains or 8.5\% relative improvement over $m_{base}$. This indicates that applying our framework can bring a net overall improvement to existing NLU. Second, analyzing per-domain results shows that $m_{new}$ outperforms $m_{base}$ (7.5-26.0\% relative improvements) on 5 domains, while making marginal improvements (0.5-3.5\% improvements) on the other 5 domains.

\paragraph{Analysis on DIM:} Table \ref{table:dim_dcm_eval}.(a) summarizes the results of error attribution annotations between the defects in the production traffic (denoted as \emph{DEF}) and target defects identified by DIM (denoted as \emph{DIM}). The results show that the target defects identified by DIM help us focus more on the defects that are caused by ASR or NLU (the ones that can be targeted and potentially fixed, specifically \emph{NLU Error} which is at 39.0\% of total for \emph{DIM} compared to 14.3\% for \emph{DEF}) and filter out others (\emph{Bad Responses} and \emph{Others}). Per-domain results show that the target defects identified by DIM consistently have a higher NLU error ratio than that of original defects for all domains.

% ------> New latex table for examples.
\begin{table*}[ht]
\aboverulesep=0ex
\belowrulesep=0ex
\centering\small
\includegraphics[width=\textwidth]{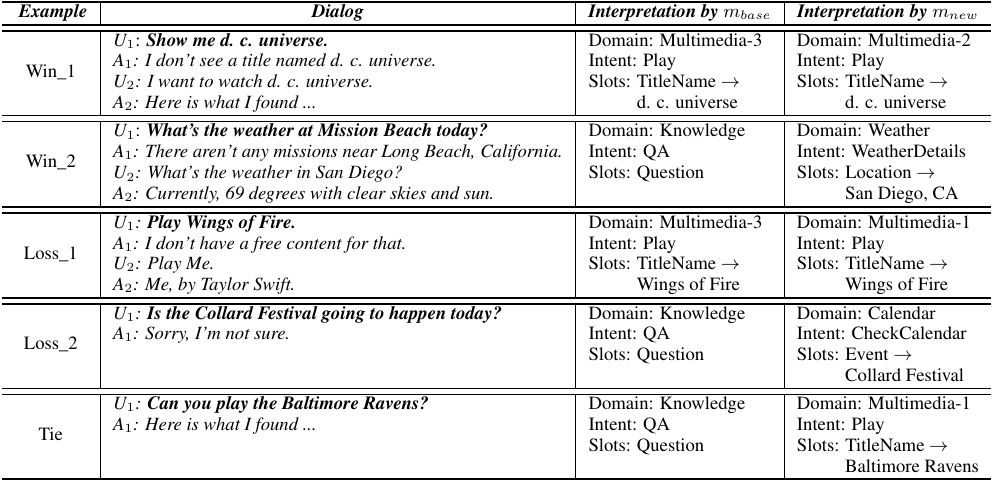}
\caption{Qualitative analysis comparing $m_{base}$ and $m_{new}$ in the overall side-by-side evaluation. For each example, the user request in bold is the turn for which the evaluation was performed. We show subsequent interaction dialog for context ($U_*$ for user requests, $A_*$ for system answers). The first two examples are ``wins" (i.e., $m_{new}$ better than $m_{base}$), followed by two ``losses" (i.e., $m_{new}$ worse than $m_{base}$), and  a ``tie" (i.e., $m_{new}$ comparable to $m_{base}$).}
\label{table:case_study}
\end{table*}

\paragraph{Analysis on DCM:} Table \ref{table:dim_dcm_eval}.(b) summarizes the win-loss annotation results on the new supervision data that take target defects from DIM and assign new interpretation labels for correction with DCM. The results show that overall DCM correctly assigns a better, corrected NLU interpretation on 399 cases and fails on 77 cases, resulting in 322 absolute gains or 32.2\% relative improvement. Per-domain results show that DCM consistently assigns a comparable or better interpretation on the target defects on almost all domains with a large margin (with 8.0\%-79.0\% relative improvements on 9 domains).

\subsection{Qualitative Analysis}
\label{subsec:case_study}
The first two examples in Table \ref{table:case_study} are wins where $m_{new}$ produced a better top interpretation than $m_{base}$. In \emph{Win 1}, $m_{base}$ produced an interpretation related to playing a title for a specific type of multimedia, while the user wanted to play the corresponding title in another multimedia type (e.g., music, video, or audio book). The updated NLU model $m_{new}$ produced the correct interpretation, most likely having learned to favor a multimedia type depending on the context, such as device status (e.g., music or video currently playing or screen is on). Similarly in \emph{Win 2}, $m_{base}$ mis-interpreted the request as a general question due to not understanding the location "Mission Beach," which is corrected by $m_{new}$.

The next two examples are losses where $m_{new}$ top-ranked incorrect interpretations such that they produced worse results than $m_{base}$. In \emph{Loss 1}, the user is in the middle of trying out a free content experience for a specific multimedia type, and we suspect the reason $m_{new}$ produced the incorrect interpretation is that there are similar requests in live traffic to "Play Wings of Fire" with another multimedia type, such that the model learns to aggressively top-rank the interpretations associated with a more dominant multimedia type. In \emph{Loss 2}, the request is for a general event query in the area, and although the Q\&A still failed to correctly answer, it was determined that it would be worse to fail in Calendar domain.

The last example is a "tie" where $m_{new}$ and $m_{base}$ both produced incorrect top interpretations that are equally bad in terms of user experience. Specifically, $m_{base}$ mis-interpreted the request as a Q\&A, while $m_{new}$ mis-interpreted the meaning of "play" for playing multimedia instead of sports. As in \emph{Loss 1}, We suspect many live utterances with the word "play" tend to be multimedia-related and biases DCM towards selecting multimedia-related interpretations.

From the qualitative analysis, especially losses, we observe that we can make our framework and new supervision data more precise if we consider more interaction history context spanning a longer period of time when we train DCM, use more signals such as personalization or subscription signals (for multimedia content types such as music or audio book). Furthermore, for truly ambiguous requests, instead of aggressively trying to correct through a new interpretation, we could offer a better experience by asking a clarifying question.

\section{Conclusion}
We proposed a domain-agnostic and scalable framework for leveraging implicit user feedback, particularly user dissatisfaction and rephrase behavior, to automatically curate new supervision data to continuously improve NLU in a large-scale conversational AI system. We showed how the framework can be applied to improve NLU and analyzed its performance across 10 popular domains on a real production system, with component-level and qualitative analysis of our framework for more in-depth validation of its performance.

\section*{Acknowledgments}
We thank Sergei Dobroshinsky, Nathan Eversole, Alex Go, Kerry Hammil, Archit Jain, Shubham Katiyar, Siddharth Mohan Misra, Joe Pemberton, and Steve Saunders for their active involvement and support for this work in the industry production system.

\bibliography{citation}

\begin{thebibliography}{43}
\expandafter\ifx\csname natexlab\endcsname\relax\def\natexlab#1{#1}\fi

\bibitem[{Beaver and Mueen(2020)}]{Beaver_Mueen_2020}
Ian Beaver and Abdullah Mueen. 2020.
\newblock Automated conversation review to surface virtual assistant
  misunderstandings: Reducing cost and increasing privacy.
\newblock In \emph{AAAI Conference on Artificial Intelligence}.

\bibitem[{Bi et~al.(2019)Bi, Teo, Dattatreya et~al.}]{bi2019leverage}
Keping Bi, Choon~Hui Teo, Yesh Dattatreya, et~al. 2019.
\newblock Leverage implicit feedback for context-aware product search.
\newblock \emph{arXiv preprint arXiv:1909.02065}.

\bibitem[{Bodigutla et~al.(2020)Bodigutla, Tiwari, Matsoukas
  et~al.}]{bodigutla2020joint}
Praveen~Kumar Bodigutla, Aditya Tiwari, Spyros Matsoukas, et~al. 2020.
\newblock Joint turn and dialogue level user satisfaction estimation on
  multi-domain conversations.
\newblock In \emph{Conference on Empirical Methods in Natural Language
  Processing}.

\bibitem[{Camacho-Collados and Pilehvar(2018)}]{camacho2018word}
Jose Camacho-Collados and Mohammad~Taher Pilehvar. 2018.
\newblock From word to sense embeddings: A survey on vector representations of
  meaning.
\newblock \emph{Journal of Artificial Intelligence Research}, 63:743--788.

\bibitem[{El-Kahky et~al.(2014)El-Kahky, Liu, Sarikaya
  et~al.}]{el2014extending}
Ali El-Kahky, Xiaohu Liu, Ruhi Sarikaya, et~al. 2014.
\newblock Extending domain coverage of language understanding systems via
  intent transfer between domains using knowledge graphs and search query click
  logs.
\newblock In \emph{International Conference on Acoustics, Speech and Signal
  Processing}.

\bibitem[{Gao et~al.(2018)Gao, Galley, and Li}]{gao2018neural}
Jianfeng Gao, Michel Galley, and Lihong Li. 2018.
\newblock Neural approaches to conversational {AI}.
\newblock In \emph{International ACM SIGIR Conference on Research and
  Development in Information Retrieval}.

\bibitem[{Hakkani-T{\"u}r et~al.(2016)Hakkani-T{\"u}r, T{\"u}r, Celikyilmaz
  et~al.}]{hakkani2016multi}
Dilek Hakkani-T{\"u}r, G{\"o}khan T{\"u}r, Asli Celikyilmaz, et~al. 2016.
\newblock Multi-domain joint semantic frame parsing using bi-directional
  {RNN-LSTM}.
\newblock In \emph{Annual Conference of the International Speech Communication
  Association}.

\bibitem[{He and McAuley(2016)}]{he2016vbpr}
Ruining He and Julian McAuley. 2016.
\newblock {VBPR}: visual bayesian personalized ranking from implicit feedback.
\newblock In \emph{AAAI Conference on Artificial Intelligence}.

\bibitem[{He et~al.(2016)He, Zhang, Kan et~al.}]{he2016fast}
Xiangnan He, Hanwang Zhang, Min-Yen Kan, et~al. 2016.
\newblock Fast matrix factorization for online recommendation with implicit
  feedback.
\newblock In \emph{International ACM SIGIR Conference on Research and
  Development in Information Retrieval}.

\bibitem[{Hu et~al.(2014)Hu, Lu, Li et~al.}]{hu2014convolutional}
Baotian Hu, Zhengdong Lu, Hang Li, et~al. 2014.
\newblock Convolutional neural network architectures for matching natural
  language sentences.
\newblock In \emph{Advances in Neural Information Processing Systems}.

\bibitem[{Hu et~al.(2008)Hu, Koren, and Volinsky}]{hu2008collaborative}
Yifan Hu, Yehuda Koren, and Chris Volinsky. 2008.
\newblock Collaborative filtering for implicit feedback datasets.
\newblock In \emph{International Conference on Data Mining}.

\bibitem[{Iyer et~al.(2017)Iyer, Konstas, Cheung et~al.}]{iyer2017learning}
Srinivasan Iyer, Ioannis Konstas, Alvin Cheung, et~al. 2017.
\newblock Learning a neural semantic parser from user feedback.
\newblock \emph{arXiv preprint arXiv:1704.08760}.

\bibitem[{Jiang et~al.(2015)Jiang, Hassan~Awadallah, Jones
  et~al.}]{jiang2015automatic}
Jiepu Jiang, Ahmed Hassan~Awadallah, Rosie Jones, et~al. 2015.
\newblock Automatic online evaluation of intelligent assistants.
\newblock In \emph{International Conference on World Wide Web}.

\bibitem[{Joachims(2002)}]{joachims2002optimizing}
Thorsten Joachims. 2002.
\newblock Optimizing search engines using clickthrough data.
\newblock In \emph{ACM SIGKDD International Conference on Knowledge Discovery
  and Data Mining}.

\bibitem[{Kepuska and Bohouta(2018)}]{kepuska2018next}
Veton Kepuska and Gamal Bohouta. 2018.
\newblock Next-generation of virtual personal assistants ({Microsoft Cortana},
  {Apple Siri}, {Amazon Alexa} and {Google Home}).
\newblock In \emph{Annual Computing and Communication Workshop and Conference}.

\bibitem[{Kingma and Ba(2014)}]{kingma2014adam}
Diederik~P Kingma and Jimmy Ba. 2014.
\newblock Adam: A method for stochastic optimization.
\newblock \emph{arXiv preprint arXiv:1412.6980}.

\bibitem[{Kiseleva et~al.(2016)Kiseleva, Williams, Jiang
  et~al.}]{kiseleva2016understanding}
Julia Kiseleva, Kyle Williams, Jiepu Jiang, et~al. 2016.
\newblock Understanding user satisfaction with intelligent assistants.
\newblock In \emph{ACM on Conference on Human Information Interaction and
  Retrieval}.

\bibitem[{Landauer et~al.(1998)Landauer, Foltz, and
  Laham}]{landauer1998introduction}
Thomas Landauer, Peter Foltz, and Darrell Laham. 1998.
\newblock An introduction to latent semantic analysis.
\newblock \emph{Discourse Processes}, 25:259--284.

\bibitem[{Li et~al.(2021)Li, Park, Dara et~al.}]{li2021neural}
Han Li, Sunghyun Park, Aswarth Dara, et~al. 2021.
\newblock Neural model robustness for skill routing in large-scale
  conversational {AI} systems: A design choice exploration.
\newblock \emph{arXiv preprint arXiv:2103.03373}.

\bibitem[{Liu and Lane(2016)}]{liu2016attention}
Bing Liu and Ian Lane. 2016.
\newblock Attention-based recurrent neural network models for joint intent
  detection and slot filling.
\newblock \emph{arXiv preprint arXiv:1609.01454}.

\bibitem[{Liu et~al.(2010)Liu, Dolan, and Pedersen}]{liu2010personalized}
Jiahui Liu, Peter Dolan, and Elin~R{\o}nby Pedersen. 2010.
\newblock Personalized news recommendation based on click behavior.
\newblock In \emph{International Conference on Intelligent User Interfaces}.

\bibitem[{Loni et~al.(2018)Loni, Larson, and Hanjalic}]{loni2018factorization}
Babak Loni, Martha Larson, and Alan Hanjalic. 2018.
\newblock Factorization machines for data with implicit feedback.
\newblock \emph{arXiv preprint arXiv:1812.08254}.

\bibitem[{Manning and Schutze(1999)}]{manning1999foundations}
Christopher Manning and Hinrich Schutze. 1999.
\newblock \emph{Foundations of Statistical Natural Language Processing}.
\newblock MIT press.

\bibitem[{Mihalcea et~al.(2006)Mihalcea, Corley, Strapparava
  et~al.}]{mihalcea2006corpus}
Rada Mihalcea, Courtney Corley, Carlo Strapparava, et~al. 2006.
\newblock Corpus-based and knowledge-based measures of text semantic
  similarity.
\newblock In \emph{AAAI Conference on Artificial Intelligence}.

\bibitem[{Mueller and Thyagarajan(2016)}]{mueller2016siamese}
Jonas Mueller and Aditya Thyagarajan. 2016.
\newblock Siamese recurrent architectures for learning sentence similarity.
\newblock In \emph{AAAI Conference on Artificial Intelligence}.

\bibitem[{Muralidharan et~al.(2019)Muralidharan, Kao, Yang
  et~al.}]{muralidharan2019leveraging}
Deepak Muralidharan, Justine Kao, Xiao Yang, et~al. 2019.
\newblock Leveraging user engagement signals for entity labeling in a virtual
  assistant.
\newblock \emph{arXiv preprint arXiv:1909.09143}.

\bibitem[{Paszke et~al.(2019)Paszke, Gross, Massa et~al.}]{paszke2019pytorch}
Adam Paszke, Sam Gross, Francisco Massa, et~al. 2019.
\newblock {PyTorch}: An imperative style, high-performance deep learning
  library.
\newblock In \emph{Advances in Neural Information Processing Systems}.

\bibitem[{Petrushkov et~al.(2018)Petrushkov, Khadivi, and
  Matusov}]{petrushkov2018learning}
Pavel Petrushkov, Shahram Khadivi, and Evgeny Matusov. 2018.
\newblock Learning from chunk-based feedback in neural machine translation.
\newblock \emph{arXiv preprint arXiv:1806.07169}.

\bibitem[{Ponnusamy et~al.(2019)Ponnusamy, Ghias, Guo et~al.}]{flare}
Pragaash Ponnusamy, Alireza~Roshan Ghias, Chenlei Guo, et~al. 2019.
\newblock Feedback-based self-learning in large-scale conversational {AI}
  agents.
\newblock \emph{arXiv preprint arXiv:1911.02557}.

\bibitem[{Reimers and Gurevych(2019)}]{reimers2019sentence}
Nils Reimers and Iryna Gurevych. 2019.
\newblock Sentence-{BERT}: Sentence embeddings using {Siamese BERT}-networks.
\newblock In \emph{Conference on Empirical Methods in Natural Language
  Processing}.

\bibitem[{Rendle et~al.(2012)Rendle, Freudenthaler, Gantner
  et~al.}]{rendle2012bpr}
Steffen Rendle, Christoph Freudenthaler, Zeno Gantner, et~al. 2012.
\newblock {BPR}: Bayesian personalized ranking from implicit feedback.
\newblock \emph{arXiv preprint arXiv:1205.2618}.

\bibitem[{Sarikaya(2017)}]{sarikaya2017technology}
Ruhi Sarikaya. 2017.
\newblock The technology behind personal digital assistants: An overview of the
  system architecture and key components.
\newblock \emph{IEEE Signal Processing Magazine}, 34:67--81.

\bibitem[{Schuster and Paliwal(1997)}]{schuster1997bidirectional}
Mike Schuster and Kuldip~K Paliwal. 1997.
\newblock Bidirectional recurrent neural networks.
\newblock \emph{IEEE Transactions on Signal Processing}, 45(11):2673--2681.

\bibitem[{Shen et~al.(2005)Shen, Tan, and Zhai}]{shen2005context}
Xuehua Shen, Bin Tan, and ChengXiang Zhai. 2005.
\newblock Context-sensitive information retrieval using implicit feedback.
\newblock In \emph{International ACM SIGIR Conference on Research and
  Development in Information Retrieval}.

\bibitem[{Srba and Bielikova(2016)}]{srba2016comprehensive}
Ivan Srba and Maria Bielikova. 2016.
\newblock A comprehensive survey and classification of approaches for community
  question answering.
\newblock \emph{Transactions on the Web}, 10:1--63.

\bibitem[{Srivastava et~al.(2015)Srivastava, Greff, and
  Schmidhuber}]{SrivastavaGS15}
Rupesh~Kumar Srivastava, Klaus Greff, and J{\"{u}}rgen Schmidhuber. 2015.
\newblock Highway networks.
\newblock \emph{arXiv preprint arXiv:1505.00387}.

\bibitem[{Su et~al.(2018)Su, Gupta, Ananthakrishnan et~al.}]{su2018re}
Chengwei Su, Rahul Gupta, Shankar Ananthakrishnan, et~al. 2018.
\newblock A re-ranker scheme for integrating large scale {NLU} models.
\newblock In \emph{Spoken Language Technology Workshop}.

\bibitem[{Sugiyama et~al.(2004)Sugiyama, Hatano, and
  Yoshikawa}]{sugiyama2004adaptive}
Kazunari Sugiyama, Kenji Hatano, and Masatoshi Yoshikawa. 2004.
\newblock Adaptive web search based on user profile constructed without any
  effort from users.
\newblock In \emph{International Conference on World Wide Web}.

\bibitem[{Ultes and Minker(2014)}]{ultes2014interaction}
Stefan Ultes and Wolfgang Minker. 2014.
\newblock Interaction quality estimation in spoken dialogue systems using
  hybrid-{HMM}s.
\newblock In \emph{Annual Meeting of the Special Interest Group on Discourse
  and Dialogue}.

\bibitem[{Wang et~al.(2021{\natexlab{a}})Wang, Kim, Park
  et~al.}]{wang2021handling}
Cheng Wang, Sun Kim, Taiwoo Park, et~al. 2021{\natexlab{a}}.
\newblock Handling long-tail queries with slice-aware conversational systems.
\newblock \emph{arXiv preprint arXiv:2104.13216}.

\bibitem[{Wang et~al.(2021{\natexlab{b}})Wang, Lee, Park
  et~al.}]{wang2021learning}
Cheng Wang, Sungjin Lee, Sunghyun Park, et~al. 2021{\natexlab{b}}.
\newblock Learning slice-aware representations with mixture of attentions.
\newblock \emph{arXiv preprint arXiv:2106.02363}.

\bibitem[{Wang et~al.(2019)Wang, Shao, and Lian}]{wang2019adversarial}
Haoyu Wang, Nan Shao, and Defu Lian. 2019.
\newblock Adversarial binary collaborative filtering for implicit feedback.
\newblock In \emph{AAAI Conference on Artificial Intelligence}.

\bibitem[{Zhang et~al.(2018)Zhang, Li, Cao et~al.}]{zhang2018exploring}
Wei-Nan Zhang, Lingzhi Li, Dongyan Cao, et~al. 2018.
\newblock Exploring implicit feedback for open domain conversation generation.
\newblock In \emph{AAAI Conference on Artificial Intelligence}.

\end{thebibliography}
\bibliographystyle{acl_natbib}

\end{document}